\documentclass[10pt,twocolumn,letterpaper]{article}

\usepackage{cvpr}
\usepackage{times}
\usepackage{epsfig}
\usepackage{graphicx}
\usepackage{amsmath}
\usepackage{amssymb}
\usepackage{booktabs}
\usepackage{booktabs}
\usepackage{stackengine}

\usepackage[normalem]{ulem}

\usepackage[pagebackref=true,breaklinks=true,letterpaper=true,colorlinks,bookmarks=false]{hyperref}

\cvprfinalcopy 

\def\cvprPaperID{687} 

\ifcvprfinal\fi
\begin{document}

\title{
Unsupervised Learning of Depth and Ego-Motion from Video}

\author{Tinghui Zhou\thanks{The majority of the work was done while interning at Google.}\\
UC Berkeley\\
\and
Matthew Brown\\
Google\\
\and
Noah Snavely\\
Google\\
\and
David G. Lowe\\
Google
}

\maketitle

\begin{abstract}
We present an unsupervised learning framework for the task of monocular depth and camera motion estimation from unstructured video sequences. In common with recent work~\cite{flynn2015deepstereo,garg2016unsupervised,godard2016unsupervised}, we use an end-to-end learning approach with view synthesis as the supervisory signal. In contrast to the previous work, our method is completely unsupervised, requiring only monocular video sequences for training. Our method uses single-view depth and multi-view pose networks, with a loss based on warping nearby views to the target using the computed depth and pose. The networks are thus coupled by the loss during training, but can be applied independently at test time. Empirical evaluation on the KITTI dataset demonstrates the effectiveness of our approach: 1) monocular depth performs comparably with supervised methods that use either ground-truth pose or depth for training, and 2) pose estimation performs favorably compared to established SLAM systems under comparable input settings.
\end{abstract}

\section{Introduction}

Humans are remarkably capable of inferring ego-motion and the 3D structure of a scene even over short timescales. For instance, in navigating along a street, we can easily locate obstacles and react quickly to avoid them. Years of research in geometric computer vision has failed to recreate similar modeling capabilities for real-world scenes (e.g., where non-rigidity, occlusion and lack of texture are present). So why do humans excel at this task? One hypothesis is that we develop a rich, structural understanding of the world through our past visual experience that has largely consisted of moving around and observing vast numbers of scenes and developing \emph{consistent} modeling of our observations. From millions of such observations, we have learned about the regularities of the world---roads are flat, buildings are straight, cars are supported by roads \etc, and we can apply this knowledge when perceiving a new scene, even from a single monocular image.

\begin{figure}[t]
    \centering
    \includegraphics[scale=0.37]{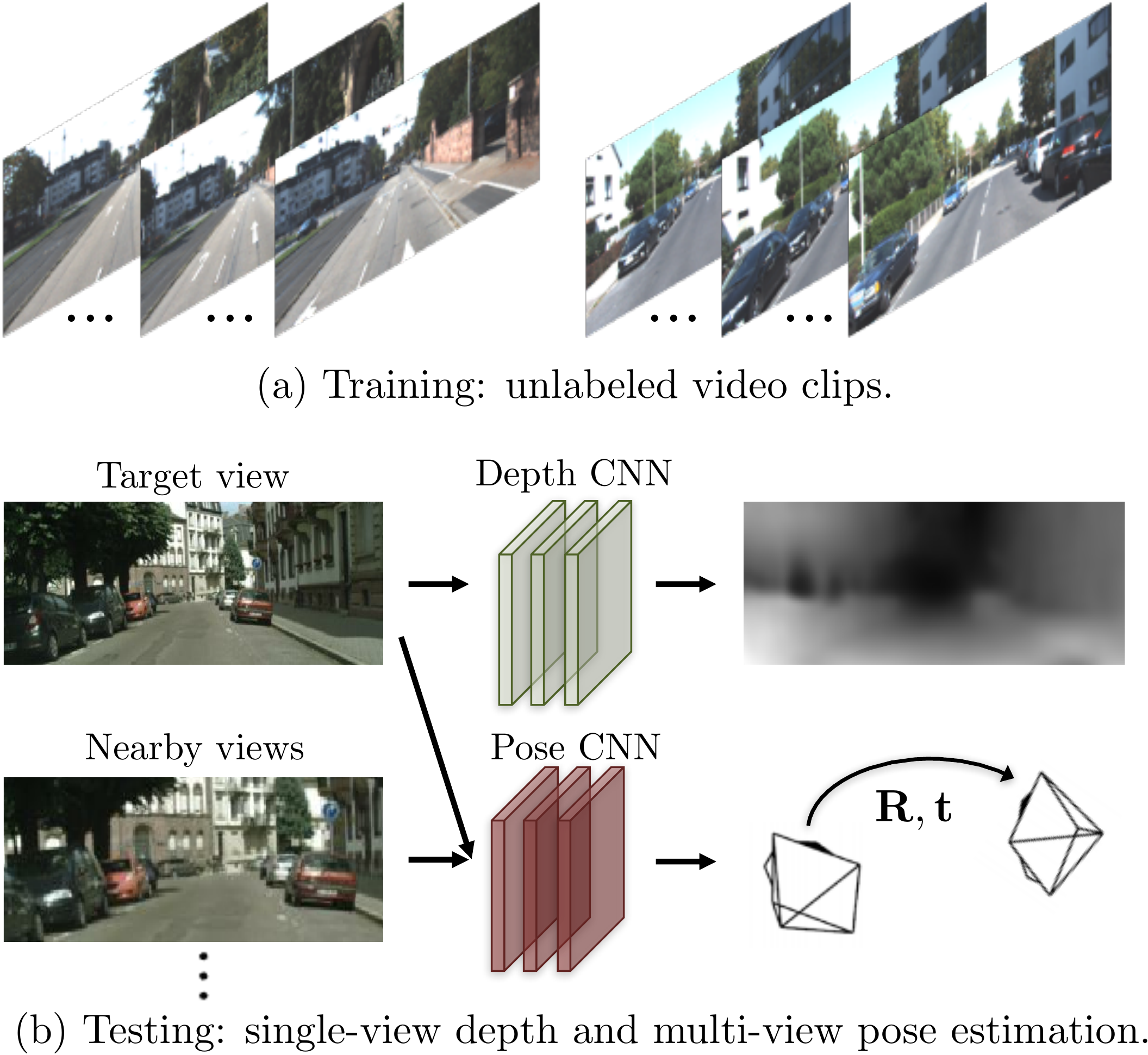}
    \caption{The training data to our system consists solely of unlabeled image sequences capturing scene appearance from different viewpoints, where the poses of the images are not provided. Our training procedure produces two models that operate independently, one for single-view depth prediction, and one for multi-view camera pose estimation.}
    \label{fig:teaser}
\end{figure}

In this work, we mimic this approach by training a model that observes sequences of images and aims to explain its observations by predicting likely camera motion and the scene structure (as shown in Fig.\ \ref{fig:teaser}). We take an end-to-end approach in allowing the model to map directly from input pixels to an estimate of ego-motion (parameterized as 6-DoF transformation matrices) and the underlying scene structure (parameterized as per-pixel depth maps under a reference view). We are particularly inspired by prior work that has suggested view synthesis as a metric~\cite{szeliski1999prediction} and recent work that tackles the calibrated, multi-view 3D case in an end-to-end framework~\cite{flynn2015deepstereo}. Our method is unsupervised, and can be trained simply using sequences of images with no manual labeling or even camera motion information. 

Our approach builds upon the insight that a geometric view synthesis system only performs \emph{consistently} well when its intermediate predictions of the scene geometry and the camera poses correspond to the physical ground-truth. While imperfect geometry and/or pose estimation can cheat with reasonable synthesized views for certain types of scenes (e.g.,\ textureless), the same model would fail miserably when presented with another set of scenes with more diverse layout and appearance structures. Thus, our goal is to formulate the entire view synthesis pipeline as the inference procedure of a convolutional neural network, so that by training the network on large-scale video data for the `meta'-task of view synthesis the network is forced to learn about intermediate tasks of depth and camera pose estimation in order to come up with a consistent explanation of the visual world. Empirical evaluation on the KITTI~\cite{geiger2012we} benchmark demonstrates the effectiveness of our approach on both single-view depth and camera pose estimation. Our code will be made available at~~\small{\url{https://github.com/tinghuiz/SfMLearner}}.

\section{Related work}

\paragraph{Structure from motion}
The simultaneous estimation of structure and motion is a well studied problem with an established toolchain of techniques~\cite{furukawa2010towards,wu2011visualsfm,newcombe2011dtam}. Whilst the traditional toolchain is effective and efficient in many cases, its reliance on accurate image correspondence can cause problems in areas of low texture, complex geometry/photometry, thin structures, and occlusions. To address these issues, several of the pipeline stages have been recently tackled using deep learning, e.g., feature matching~\cite{han2015matchnet}, pose estimation~\cite{kendall2015posenet}, and stereo~\cite{flynn2015deepstereo,kendall2017end,zbontar2016stereo}. These learning-based techniques are attractive in that they are able to leverage external supervision during training, and potentially overcome the above issues when applied to test data.

\paragraph{Warping-based view synthesis}
One important application of geometric scene understanding is the task of novel view synthesis, where the goal is to synthesize the appearance of the scene seen from novel camera viewpoints. A classic paradigm for view synthesis is to first either estimate the underlying 3D geometry explicitly or establish pixel correspondence among input views, and then synthesize the novel views by compositing image patches from the input views (e.g.,~\cite{chen1993view,zitnick2004high,seitz1996view,debevec1996modeling,fitzgibbon2005image}). Recently, end-to-end learning has been applied to reconstruct novel views by transforming the input based on depth or flow, e.g., DeepStereo~\cite{flynn2015deepstereo}, Deep3D~\cite{xie2016deep} and Appearance Flows~\cite{zhou2016view}. In these methods, the underlying geometry is represented by quantized depth planes (DeepStereo), probabilistic disparity maps (Deep3D) and view-dependent flow fields (Appearance Flows), respectively. Unlike methods that directly map from input views to the target view (e.g.,~\cite{tatarchenko2016multi}), warping-based methods are forced to learn intermediate predictions of geometry and/or correspondence. In this work, we aim to distill such geometric reasoning capability from CNNs trained to perform warping-based view synthesis.


\paragraph{Learning single-view 3D from registered 2D views}
Our work is closely related to a line of recent research on learning single-view 3D inference from registered 2D observations. Garg~\etal~\cite{garg2016unsupervised} propose to learn a single-view depth estimation CNN using projection errors to a calibrated stereo twin for supervision. Concurrently, Deep3D~\cite{xie2016deep} predicts a second stereo viewpoint from an input image using stereoscopic film footage as training data. A similar approach was taken by Godard~\etal~\cite{godard2016unsupervised}, with the addition of a left-right consistency constraint, and a better architecture design that led to impressive performance. Like our approach, these techniques only learn from image observations of the world, unlike methods that require explicit depth for training, e.g.,~\cite{hoiem2005photo,saxena2009make3d,eigen2014depth,kendall2017end,kuznietsov2017semi}. 

These techniques bear some resemblance to direct methods for structure and motion estimation~\cite{irani1999direct}, where the camera parameters and scene depth are adjusted to minimize a pixel-based error function. However, rather than directly minimizing the error to obtain the estimation, the CNN-based methods only take a gradient step for each batch of input instances, which allows the network to learn an implicit prior from a large corpus of related imagery. Several authors have explored building differentiable rendering operations into their models that are trained in this way, e.g.,~\cite{handa2016gvnn,kulkarni2015,loper2014opendr}. 

While most of the above techniques (including ours) are mainly focused on inferring depth maps as the scene geometry output, recent work (e.g.,~\cite{gadelha20163d,rezende2016unsupervised,tulsiani2017multi,yan2016perspective}) has also shown success in learning 3D volumetric representations from 2D observations based on similar principles of projective geometry.  Fouhey~\etal~\cite{fouhey2015single} further show that it is even possible to learn 3D inference without 3D labels (or registered 2D views) by utilizing scene regularity.

\paragraph{Unsupervised/Self-supervised learning from video}
Another line of related work to ours is visual representation learning from video, where the general goal is to design pretext tasks for learning generic visual features from video data that can later be re-purposed for other vision tasks such as object detection and semantic segmentation. Such pretext tasks include ego-motion estimation~\cite{agrawal2015learning,jayaraman-iccv2015}, tracking~\cite{wang2015unsupervised}, temporal coherence~\cite{goroshin2015unsupervised}, temporal order verification~\cite{misra2016shuffle}, and object motion mask prediction~\cite{pathakCVPR17learning}. While we focus on inferring the explicit scene geometry and ego-motion in this work, intuitively, the internal representation learned by the deep network (especially the single-view depth CNN) should capture some level of semantics that could generalize to other tasks as well.

Concurrent to our work, Vijayanarasimhan~\etal~\cite{Vijayanarasimhan2017SfM} independently propose a framework for joint training of depth, camera motion and scene motion from videos. While both methods are conceptually similar, ours is focused on the unsupervised aspect, whereas their framework adds the capability to incorporate supervision (e.g., depth, camera motion or scene motion). There are significant differences in how scene dynamics are modeled during training, in which they explicitly solve for object motion whereas our explainability mask discounts regions undergoing motion, occlusion and other factors.

\section{Approach}
Here we propose a framework for jointly training a single-view depth CNN and a camera pose estimation CNN from unlabeled video sequences. Despite being jointly trained, the depth model and the pose estimation model can be used independently during test-time inference. Training examples to our model consist of short image sequences of scenes captured by a moving camera. While our training procedure is robust to some degree of scene motion, we assume that the scenes we are interested in are mostly rigid, i.e., the scene appearance change across different frames is dominated by the camera motion. 

\begin{figure}[t]
\centering
\includegraphics[width=\linewidth]{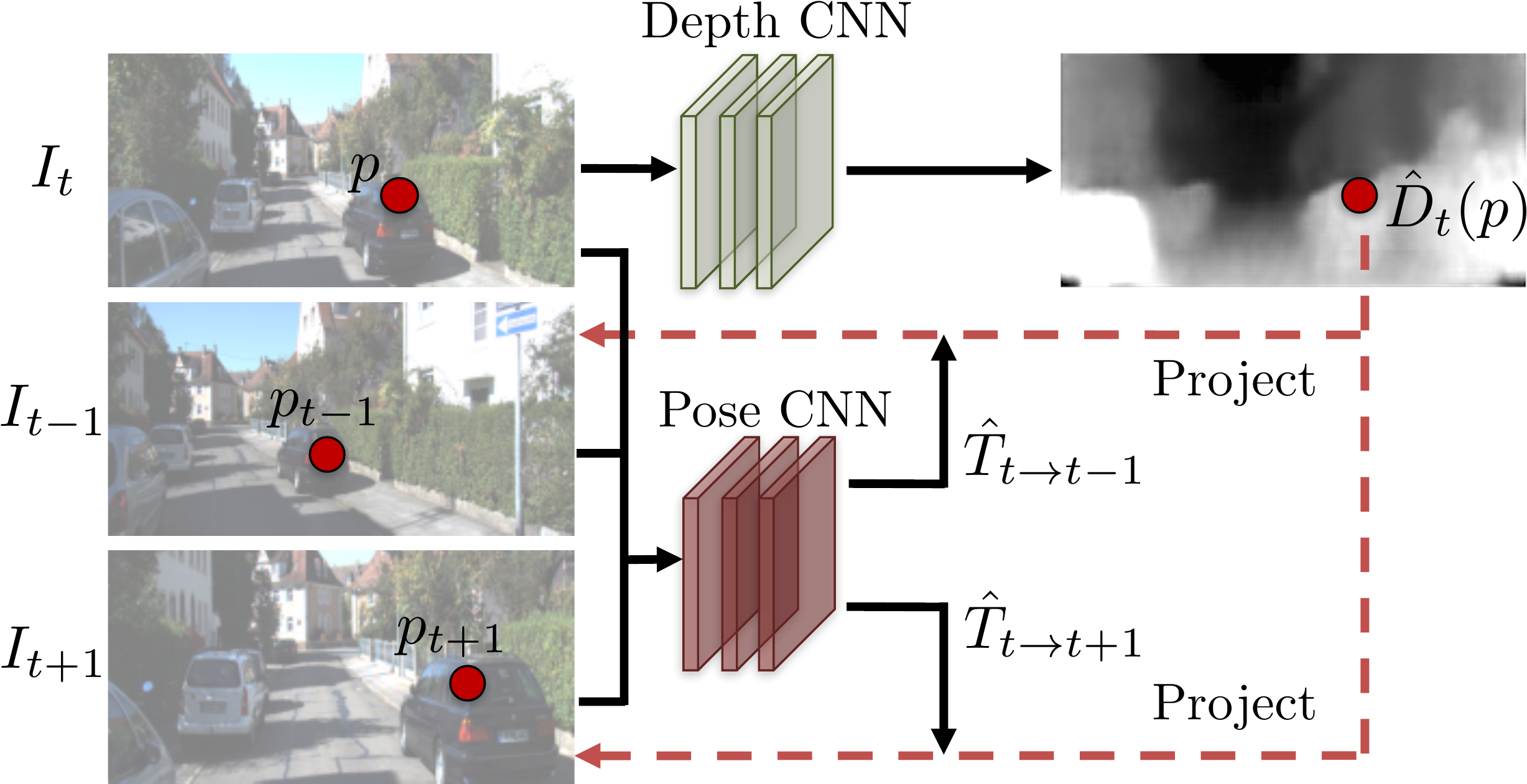}
\caption{ Overview of the supervision pipeline based on view synthesis. The depth network takes only the target view as input, and outputs a per-pixel depth map $\hat{D}_t$. The pose network takes both the target view ($I_t$) and the nearby/source views (e.g., $I_{t-1}$ and $I_{t+1}$) as input, and outputs the relative camera poses ($\hat{T}_{t\rightarrow t-1}, \hat{T}_{t\rightarrow t+1}$). The outputs of both networks are then used to inverse warp the source views (see Sec.~\ref{sec:proj}) to reconstruct the target view, and the photometric reconstruction loss is used for training the CNNs. By utilizing view synthesis as supervision, we are able to train the entire framework in an unsupervised manner from videos.}
\label{fig:method}
\end{figure}

\subsection{View synthesis as supervision}
The key supervision signal for our depth and pose prediction CNNs comes from the task of \emph{novel view synthesis}: given one input view of a scene, synthesize a new image of the scene seen from a different camera pose. We can synthesize a target view given a per-pixel depth in that image, plus the pose and visibility in a nearby view. As we will show next, this synthesis process can be implemented in a fully differentiable manner with CNNs as the geometry and pose estimation modules. Visibility can be handled, along with non-rigidity and other non-modeled factors, using an ``explanability" mask, which we discuss later (Sec.~\ref{sec:explanability}).

Let us denote $<I_1, \ldots, I_N>$ as a training image sequence with one of the frames $I_t$ being the target view and the rest being the source views $I_s (1 \le s \le N, s \neq t)$. The view synthesis objective can be formulated as
\begin{equation}
    \mathcal{L}_{vs} = \sum_s\sum_p | I_t (p) - \hat{I}_s(p) |~,
    \label{eq:obj}
\end{equation}
where $p$ indexes over pixel coordinates, and $\hat{I}_s$ is the source view $I_s$ warped to the target coordinate frame based on a depth image-based rendering module~\cite{fehn2004depth} (described in Sec.~\ref{sec:proj}), taking the predicted depth $\hat{D}_t$, the predicted $4\times 4$ camera transformation matrix\footnote{In practice, the CNN estimates the Euler angles and the 3D translation vector, which are then converted to the transformation matrix.} $\hat{T}_{t\rightarrow s}$ and the source view $I_s$  as input.  

\begin{figure}[t]
\centering
\includegraphics[width=\linewidth]{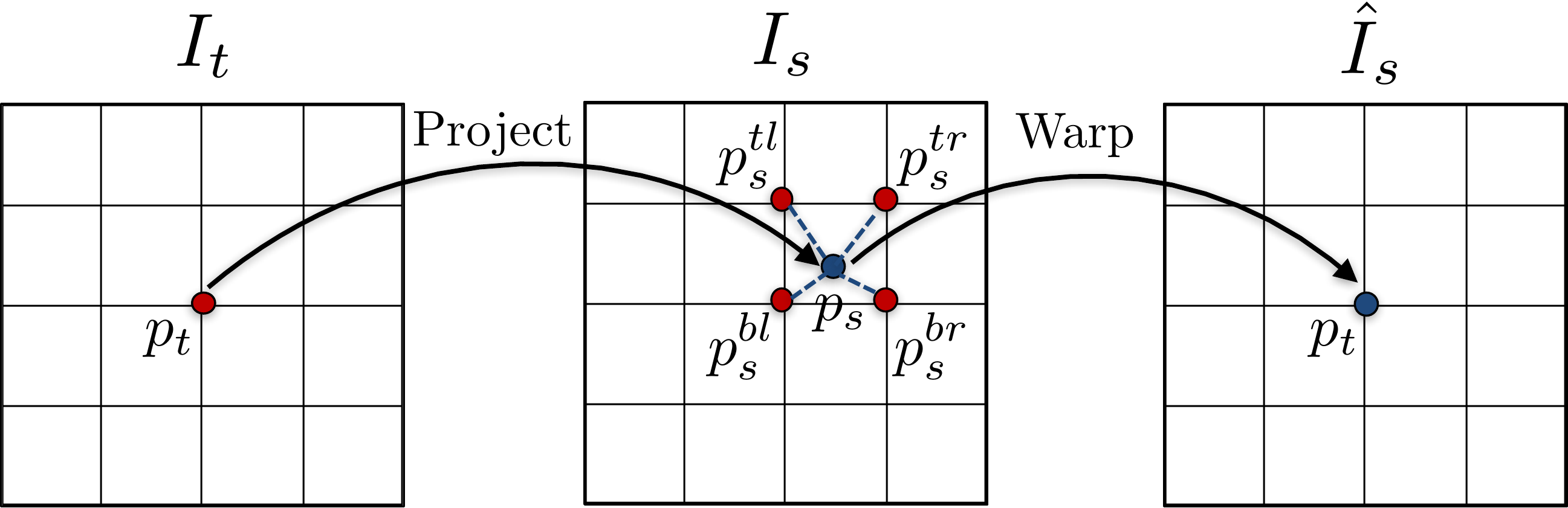}
\caption{Illustration of the differentiable image warping process. For each point $p_t$ in the target view, we first project it onto the source view based on the predicted depth and camera pose, and then use bilinear interpolation to obtain the value of the warped image $\hat{I}_s$ at location $p_t$.}
\label{fig:warp}
\end{figure}

Note that the idea of view synthesis as supervision has also been recently explored for learning single-view depth estimation~\cite{garg2016unsupervised,godard2016unsupervised} and multi-view stereo~\cite{flynn2015deepstereo}. However, to the best of our knowledge, all previous work requires posed image sets during training (and testing too in the case of DeepStereo), while our framework can be applied to standard videos without pose information. Furthermore, it predicts the poses as part of the learning framework. See Figure~\ref{fig:method} for an illustration of our learning pipeline for depth and pose estimation.

\subsection{Differentiable depth image-based rendering}
\label{sec:proj}
As indicated in Eq.~\ref{eq:obj}, a key component of our learning framework is a differentiable depth image-based renderer that reconstructs the target view $I_t$ by sampling pixels from a source view $I_s$ based on the predicted depth map $\hat{D}_t$ and the relative pose $\hat{T}_{t \rightarrow s}$. 

 Let $p_t$ denote the homogeneous coordinates of a pixel in the target view, and $K$ denote the camera intrinsics matrix. We can obtain $p_t$'s projected coordinates onto the source view $p_s$ by\footnote{For notation simplicity, we omit showing the necessary conversion to homogeneous coordinates along the steps of matrix multiplication.}
\begin{equation}
    p_s \sim K \hat{T}_{t\rightarrow s}\hat{D}_{t}(p_t)K^{-1}p_t
    \label{eq:proj}
\end{equation}
Notice that the projected coordinates $p_s$ are continuous values. To obtain $I_s(p_s)$ for populating the value of $\hat{I}_s(p_t)$ (see Figure~\ref{fig:warp}), we then use the differentiable bilinear sampling mechanism proposed in the \emph{spatial transformer networks}~\cite{jaderberg2015spatial} that linearly interpolates the values of the $4$-pixel neighbors (top-left, top-right, bottom-left, and bottom-right) of $p_s$ to approximate $I_s(p_s)$, i.e. $\hat{I}_s(p_t) = I_s(p_s) = \sum_{i\in\{t,b\}, j\in\{l,r\}}w^{ij}I_s(p_s^{ij}),$ where $w^{ij}$ is linearly proportional to the spatial proximity between $p_s$ and $p_s^{ij}$ , and $\sum_{i,j}w^{ij} = 1$. A similar strategy is used in~\cite{zhou2016view} for learning to directly warp between different views, while here the coordinates for pixel warping are obtained through projective geometry that enables the factorization of depth and camera pose.

\begin{figure*}[t]
    \centering
    \includegraphics[width=0.98\linewidth]{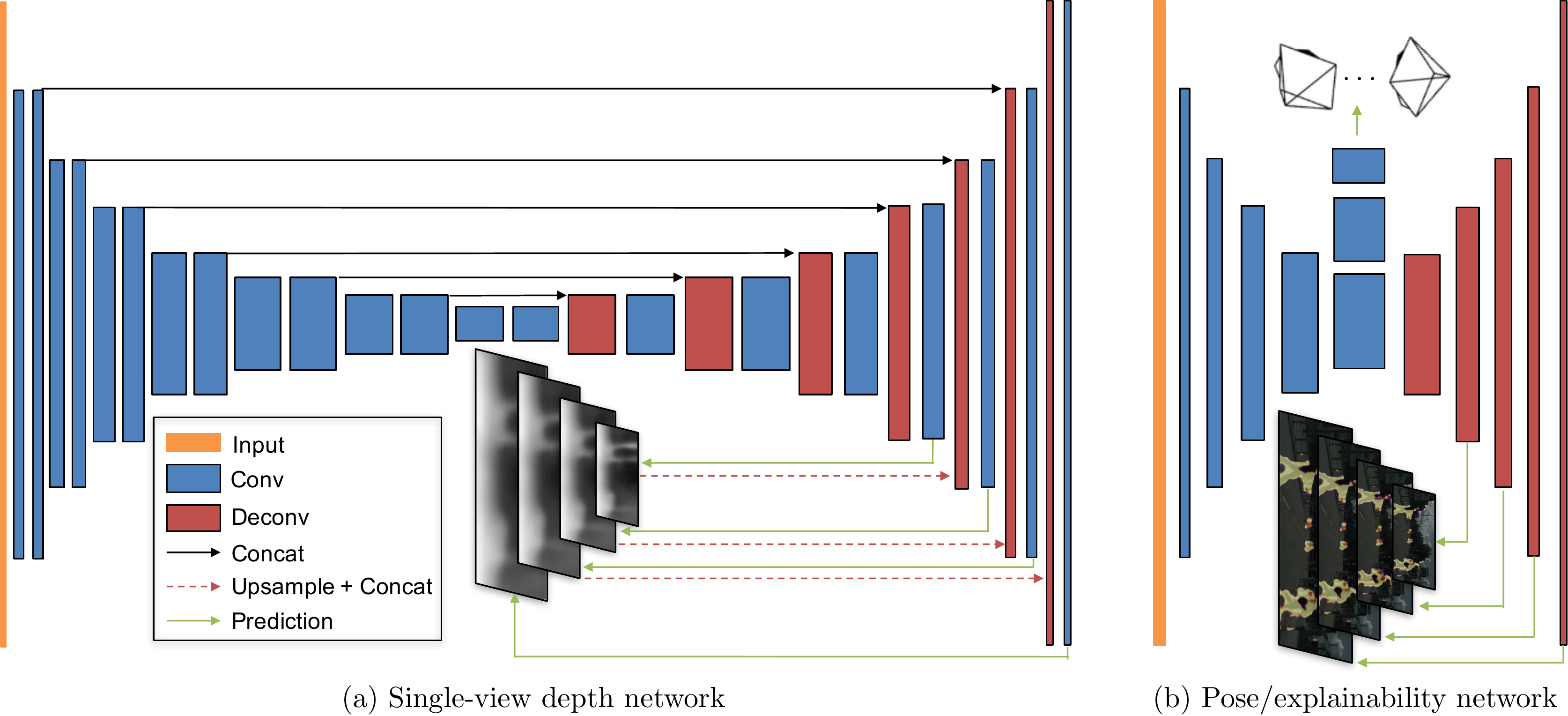}
    \caption{Network architecture for our depth/pose/explainability prediction modules. The width and height of each rectangular block indicates the output channels and the spatial dimension of the feature map at the corresponding layer respectively, and each reduction/increase in size indicates a change by the factor of $2$. (a) For single-view depth, we adopt the DispNet~\cite{mayer2016large} architecture with multi-scale side predictions. The kernel size is $3$ for all the layers except for the first $4$ conv layers with $7, 7, 5, 5$, respectively. The number of output channels for the first conv layer is $32$. (b) The pose and explainabilty networks share the first few conv layers, and then branch out to predict 6-DoF relative pose and multi-scale explainability masks, respectively. The number of output channels for the first conv layer is $16$, and the kernel size is $3$ for all the layers except for the first two conv and the last two deconv/prediction layers where we use $7, 5, 5, 7$, respectively.  See Section~\ref{sec:net} for more details.}
    \label{fig:net}
\end{figure*}

\subsection{Modeling the model limitation}
\label{sec:explanability}
Note that when applied to monocular videos the above view synthesis formulation implicitly assumes 1) the scene is static without moving objects; 2) there is no occlusion/disocclusion between the target view and the source views; 3) the surface is Lambertian so that the photo-consistency error is meaningful. If any of these assumptions are violated in a training sequence, the gradients could be corrupted and potentially inhibit training. To improve the robustness of our learning pipeline to these factors, we additionally train a \emph{explainability prediction} network (jointly and simultaneously with the depth and pose networks) that outputs a per-pixel soft mask $\hat{E}_s$ for each target-source pair, indicating the network's belief in where direct view synthesis will be successfully modeled for each target pixel. Based on the predicted $\hat{E}_s$, the view synthesis objective is weighted correspondingly by
\begin{equation}
    \mathcal{L}_{vs} = \sum_{<I_1, \ldots, I_N> \in \mathcal{S}}\sum_p \hat{E}_s(p) | I_t (p) - \hat{I}_s(p) |~.
\label{eq:exp}
\end{equation}
Since we do not have direct supervision for $\hat{E}_s$, training with the above loss would result in a trivial solution of the network always predicting $\hat{E}_s$ to be zero, which perfectly minimizes the loss. To resolve this, we add a regularization term $\mathcal{L}_{reg}(\hat{E}_s)$ that encourages nonzero predictions by minimizing the cross-entropy loss with constant label $1$ at each pixel location. In other words, the network is encouraged to minimize the view synthesis objective, but allowed a certain amount of slack for discounting the factors not considered by the model. 

\subsection{Overcoming the gradient locality}
One remaining issue with the above learning pipeline is that the gradients are mainly derived from the pixel intensity difference between $I(p_t)$ and the four neighbors of $I(p_s)$, which would inhibit training if the correct $p_s$ (projected using the ground-truth depth and pose) is located in a low-texture region or far from the current estimation. This is a well known issue in motion estimation~\cite{bergen1992hierarchical}. Empirically, we found two strategies to be effective for overcoming this issue: 1) using a convolutional encoder-decoder architecture with a small bottleneck for the depth network that implicitly constrains the output to be globally smooth and facilitates gradients to propagate from meaningful regions to nearby regions; 2) explicit multi-scale and smoothness loss (e.g., as in~\cite{garg2016unsupervised,godard2016unsupervised}) that allows gradients to be derived from larger spatial regions directly. We adopt the second strategy in this work as it is less sensitive to architectural choices. For smoothness, we minimize the $L_1$ norm of the second-order gradients for the predicted depth maps (similar to~\cite{Vijayanarasimhan2017SfM}).

Our final objective becomes
\begin{equation}
    \mathcal{L}_{final} = \sum_l \mathcal{L}_{vs}^l +  \lambda_s \mathcal{L}^l_{smooth} + \lambda_e \sum_{s} \mathcal{L}_{reg}(\hat{E}_s^l)~,
\end{equation}
where $l$ indexes over different image scales, $s$ indexes over source images, and $\lambda_s$ and $\lambda_e$ are the weighting for the depth smoothness loss and the explainability regularization, respectively.

\subsection{Network architecture}
\label{sec:net}

\paragraph{Single-view depth}
For single-view depth prediction, we adopt the DispNet architecture proposed in~\cite{mayer2016large} that is mainly based on an encoder-decoder design with skip connections and multi-scale side predictions (see Figure~\ref{fig:net}). All conv layers are followed by ReLU activation except for the prediction layers, where we use $1/(\alpha * sigmoid(x) + \beta)$ with $\alpha = 10$ and $\beta = 0.01$ to constrain the predicted depth to be always positive within a reasonable range. We also experimented with using multiple views as input to the depth network, but did not find this to improve the results. This is in line with the observations in~\cite{ummenhofer2016demon}, where optical flow constraints need to be enforced to utilize multiple views effectively.

\paragraph{Pose}
The input to the pose estimation network is the target view concatenated with all the source views (along the color channels), and the outputs are the relative poses between the target view and each of the source views. The network consists of $7$ stride-2 convolutions followed by a $1\times 1$ convolution with $6 * (N-1)$ output channels (corresponding to $3$ Euler angles and $3$-D translation for each source view). Finally, global average pooling is applied to aggregate predictions at all spatial locations. All conv layers are followed by ReLU except for the last layer where no nonlinear activation is applied.

\paragraph{Explainability mask}
The explainability prediction network shares the first five feature encoding layers with the pose network, followed by $5$ deconvolution layers with multi-scale side predictions. All conv/deconv layers are followed by ReLU except for the prediction layers with no nonlinear activation. The number of output channels for each prediction layer is $2 * (N-1)$, with every two channels normalized by \emph{softmax} to obtain the explainability prediction for the corresponding source-target pair (the second channel after normalization is $\hat{E}_s$ and used in computing the loss in Eq.~\ref{eq:exp}).  

\section{Experiments}
Here we evaluate the performance of our system, and compare with prior approaches on single-view depth as well as ego-motion estimation. We mainly use the KITTI dataset~\cite{geiger2012we} for benchmarking, but also use the Make3D dataset~\cite{saxena2009make3d} for evaluating cross-dataset generalization ability.

\paragraph{Training Details} We implemented the system using the publicly available TensorFlow~\cite{abadi2016tensorflow} framework. For all the experiments, we set $\lambda_s=0.5/l$ ($l$ is the downscaling factor for the corresponding scale) and $\lambda_e = 0.2$. During training, we used batch normalization~\cite{ioffe2015batch} for all the layers except for the output layers, and the Adam~\cite{kingma2014adam} optimizer with $\beta_1 = 0.9$, $\beta_2 = 0.999$, learning rate of $0.0002$ and mini-batch size of $4$. The training typically converges after about $150K$ iterations. All the experiments are performed with image sequences captured with a monocular camera. We resize the images to $128\times 416$ during training, but both the depth and pose networks can be run fully-convolutionally for images of arbitrary size at test time.

\begin{figure}[t]
    \centering
    \includegraphics[width=1.0\linewidth]{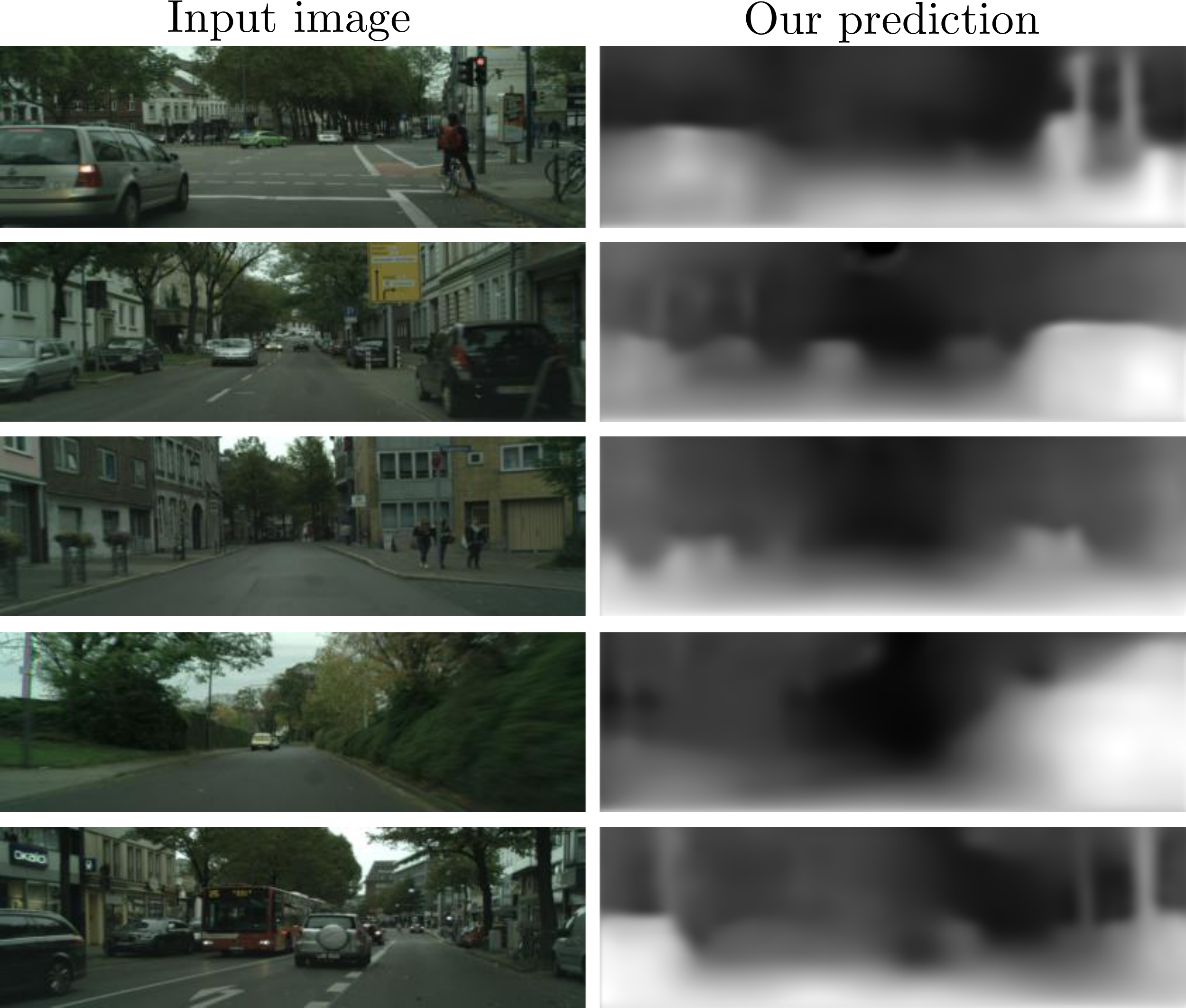}
    \caption{Our sample predictions on the Cityscapes dataset using the model trained on Cityscapes only.}
    \label{fig:city}
\end{figure}

\begin{figure*}[t]
\centering
\includegraphics[width=\linewidth]{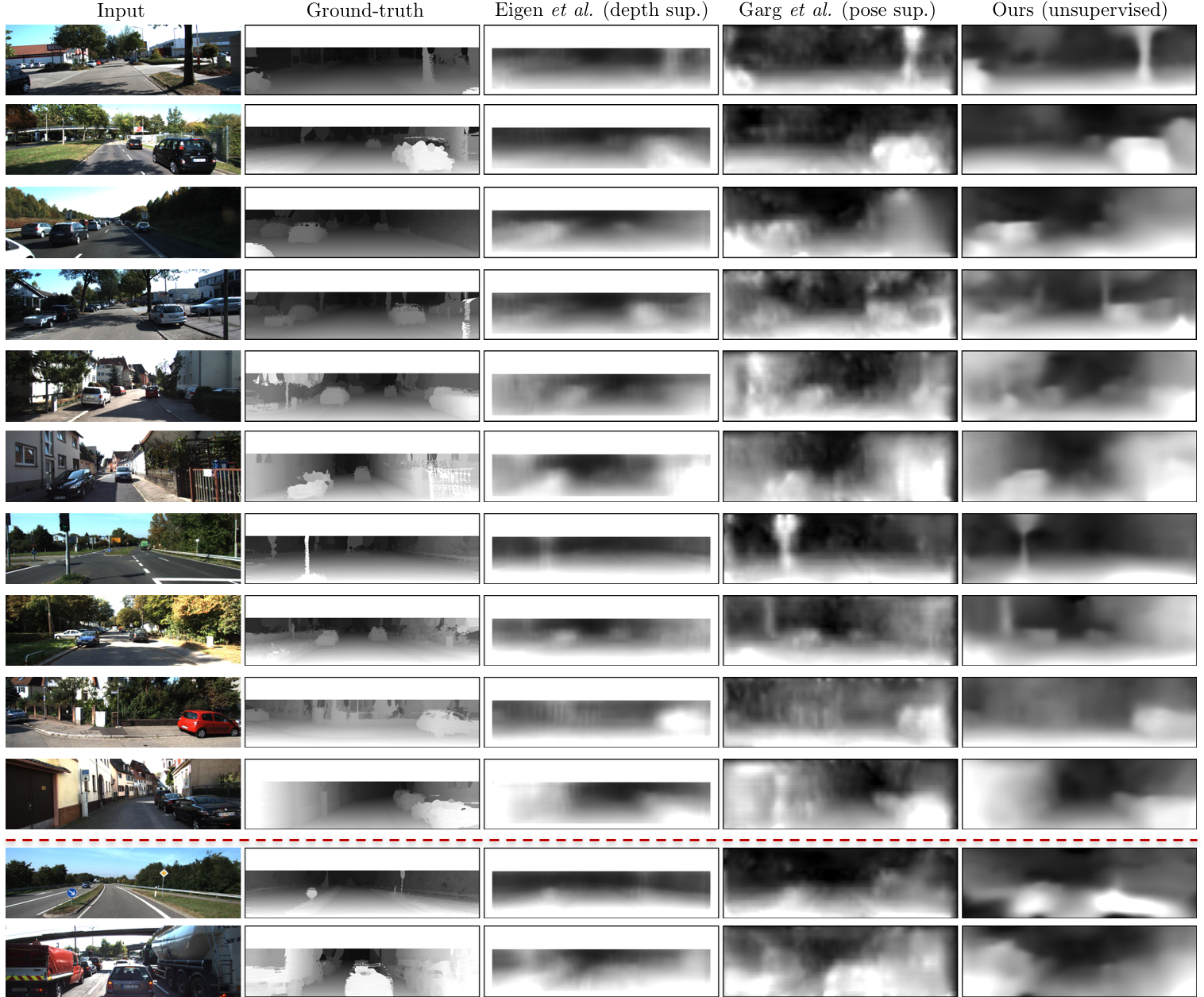}
\caption{Comparison of single-view depth estimation between Eigen~\etal~\cite{eigen2014depth} (with ground-truth depth supervision), Garg~\etal~\cite{garg2016unsupervised} (with ground-truth pose supervision), and ours (unsupervised). The ground-truth depth map is interpolated from sparse measurements for visualization purpose. The last two rows show typical failure cases of our model, which sometimes struggles in vast open scenes and objects close to the front of the camera.}
\label{fig:mono_res}
\end{figure*}

\begin{figure}[t]
    \centering
    \includegraphics[width=1.0\linewidth]{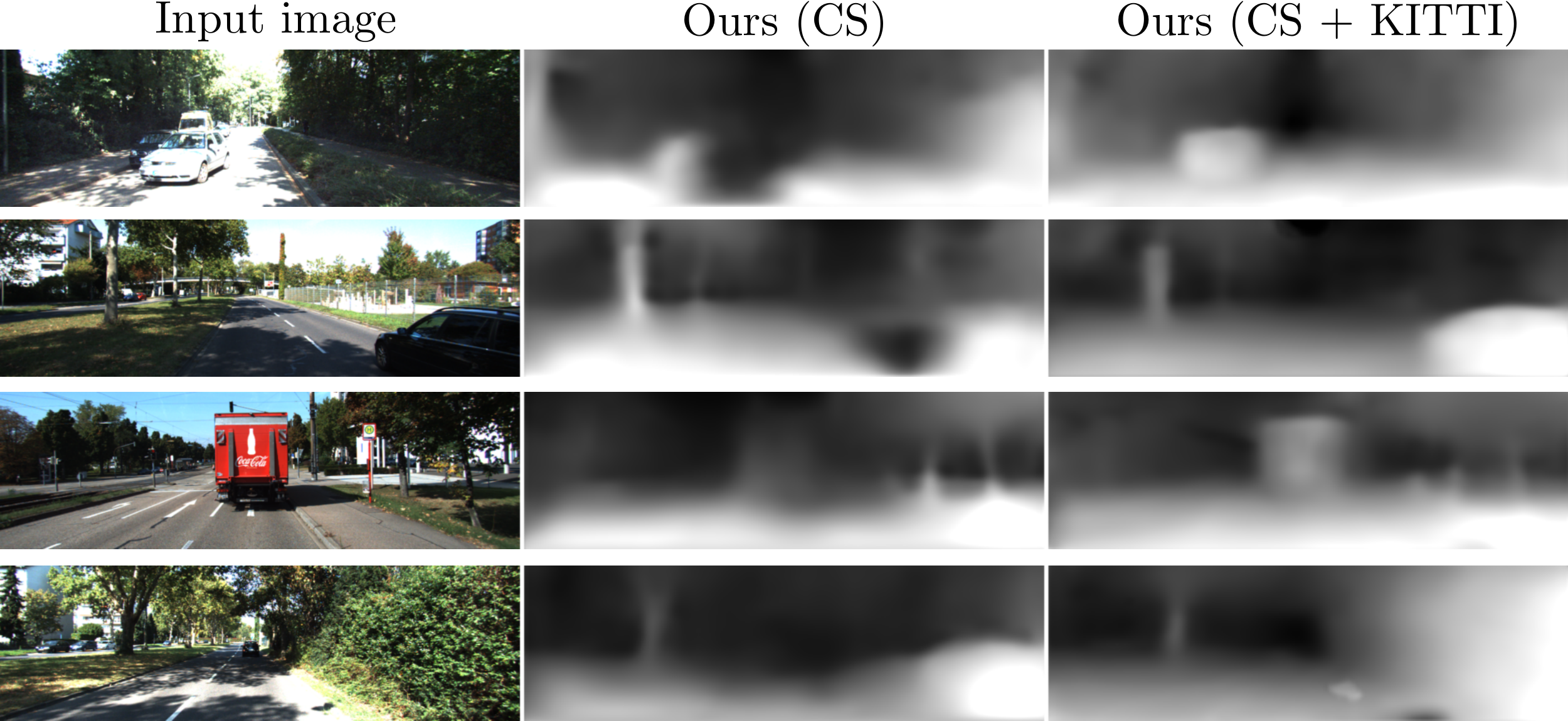}
    \caption{Comparison of single-view depth predictions on the KITTI dataset by our initial Cityscapes model and the final model (pre-trained on Cityscapes and then fine-tuned on KITTI). The Cityscapes model sometimes makes structural mistakes (e.g. holes on car body) likely due to the domain gap between the two datasets.}
    \label{fig:kitti_ft}
\end{figure}

\subsection{Single-view depth estimation}
We train our system on the split provided by~\cite{eigen2014depth}, and exclude all the frames from the testing scenes as well as static sequences with mean optical flow magnitude less than $1$ pixel for training. We fix the length of image sequences to be $3$ frames, and treat the central frame as the target view and the $\pm 1$ frames as the source views. We use images captured by both color cameras, but treated them independently when forming training sequences. This results in a total of $44,540$ sequences, out of which we use $40,109$ for training and $4,431$ for validation.

To the best of our knowledge, no previous systems exist that learn single-view depth estimation in an unsupervised manner from monocular videos. Nonetheless, here we provide comparison with prior methods with depth supervision~\cite{eigen2014depth} and recent methods that use calibrated stereo images (i.e. with pose supervision) for training~\cite{garg2016unsupervised,godard2016unsupervised}.
Since the depth predicted by our method is defined up to a scale factor, for evaluation we multiply the predicted depth maps by a scalar $\hat{s}$ that matches the median with the ground-truth, i.e. $\hat{s} = median(D_{gt})/median(D_{pred})$. 

Similar to~\cite{godard2016unsupervised}, we also experimented with first pre-training the system on the larger Cityscapes dataset~\cite{cordts2016cityscapes} (sample predictions are shown in Figure~\ref{fig:city}), and then fine-tune on KITTI, which results in slight performance improvement.  


\begin{table*}[t]
  \centering
  \resizebox{1.0\textwidth}{!}{
  \begin{tabular}{lcccccccccc}
  \toprule
  Method & Dataset & \multicolumn{2}{c}{Supervision}  & \multicolumn{4}{c}{Error metric} & \multicolumn{3}{c}{Accuracy metric}\\
  \cmidrule(lr){3-4}
  \cmidrule(lr){5-8}
  \cmidrule(lr){9-11}
   & & Depth & Pose  & Abs Rel & Sq Rel & RMSE  & RMSE log & $\delta < 1.25 $ & $\delta < 1.25^{2}$ & $\delta < 1.25^{3}$\tabularnewline
  \midrule
  Train set mean & K & \checkmark &  & 0.403 & 5.530 & 8.709 & 0.403 & 0.593 & 0.776 & 0.878 \tabularnewline    
  Eigen \etal~\cite{eigen2014depth} Coarse & K & \checkmark  & & 0.214 & 1.605 & 6.563 & 0.292 & 0.673 & 0.884 & 0.957 \tabularnewline 
  Eigen \etal~\cite{eigen2014depth} Fine & K & \checkmark & & 0.203 & 1.548 & 6.307 & 0.282 & 0.702 & 0.890 & 0.958 \tabularnewline 
  Liu \etal~\cite{liu2016learning} & K & \checkmark & & 0.202 & 1.614 & 6.523 & 0.275 & 0.678 & 0.895 & 0.965 \tabularnewline
  Godard~\etal~\cite{godard2016unsupervised} & K & & \checkmark& 0.148 & 1.344 & 5.927 & 0.247	& 0.803 & 0.922 & 0.964 \tabularnewline
  Godard~\etal~\cite{godard2016unsupervised} & CS + K & & \checkmark & 0.124 & 1.076 & 5.311 & 0.219	& 0.847 & 0.942 & 0.973 \tabularnewline
  \textbf{Ours} (w/o explainability) & K &  & & 0.221 & 2.226 & 7.527 & 0.294 &     0.676 & 0.885 & 0.954 \tabularnewline
  \textbf{Ours} & K &  & & 0.208 & 1.768 & 6.856 & 0.283 &  0.678 &  0.885 & 0.957 \tabularnewline
  \textbf{Ours} & CS & & & 0.267 &  2.686 & 7.580 & 0.334 &     0.577 & 0.840 & 0.937 \tabularnewline
  \textbf{Ours} & CS + K & & & 0.198 & 1.836 & 6.565 & 0.275 & 0.718 & 0.901 & 0.960 \tabularnewline
  \midrule
  Garg \etal~\cite{garg2016unsupervised} cap 50m & K & & \checkmark & 0.169 & 1.080 & 5.104 & 0.273 & 0.740 & 0.904 & 0.962 \tabularnewline
  \textbf{Ours} (w/o explainability) cap 50m & K & & & 0.208  & 1.551 &  5.452  & 0.273 & 0.695 & 0.900 & 0.964 \tabularnewline
  \textbf{Ours} cap 50m & K & & & 0.201 & 1.391 & 5.181 &     0.264 & 0.696 & 0.900 & 0.966 \tabularnewline
  \textbf{Ours} cap 50m & CS & & &  0.260 &  2.232 & 6.148 &     0.321 & 0.590 & 0.852 & 0.945 \tabularnewline
  \textbf{Ours} cap 50m &  CS + K & & & 0.190 & 1.436 & 4.975 & 0.258 & 0.735 & 0.915 & 0.968 \tabularnewline
  \bottomrule
  \end{tabular}
  }
  \vspace{10pt}
  \caption{Single-view depth results on the KITTI dataset \cite{geiger2012we} using the split of Eigen~\etal~\cite{eigen2014depth} (Baseline numbers taken from~\cite{godard2016unsupervised}). For training, K = KITTI, and CS = Cityscapes \cite{cordts2016cityscapes}. All methods we compare with use some form of supervision (either ground-truth depth or calibrated camera pose) during training. Note: results from Garg et al.~\cite{garg2016unsupervised} are capped at 50m depth, so we break these out separately in the lower part of the table.}
    \label{tab:kitti_eigen}
    \vspace{-10pt}
\end{table*}


\paragraph{KITTI}
Here we evaluate the single-view depth performance on the $697$ images from the test split of~\cite{eigen2014depth}. As shown in Table~\ref{tab:kitti_eigen}, our unsupervised method performs comparably with several supervised methods (e.g. Eigen~\etal~\cite{eigen2014depth} and Garg~\etal~\cite{garg2016unsupervised}), but falls short of concurrent work by Godard~\etal~\cite{godard2016unsupervised} that uses calibrated stereo images (i.e. with pose supervision) with left-right cycle consistency loss for training. For future work, it would be interesting to see if incorporating the similar cycle consistency loss into our framework could further improve the results. Figure~\ref{fig:mono_res} provides examples of visual comparison between our results and some supervised baselines over a variety of examples. One can see that although trained in an unsupervised manner, our results are comparable to that of the supervised baselines, and sometimes preserve the depth boundaries and thin structures such as trees and street lights better. 

We show sample predictions made by our initial Cityscapes model and the final model (pre-trained on Cityscapes and then fine-tuned on KITTI) in Figure~\ref{fig:kitti_ft}. Due to the domain gap between the two datasets, our Cityscapes model sometimes has difficulty in recovering the complete shape of the car/bushes, and mistakes them with distant objects.

We also performed an ablation study of the explainability modeling~(see Table~\ref{tab:kitti_eigen}), which turns out only offering a modest performance boost. This is likely because 1) most of the KITTI scenes are static without significant scene motions, and 2) the occlusion/visibility effects only occur in small regions in sequences across a short time span ($3$-frames), which make the explainability modeling less essential to the success of training. Nonetheless, our explainability prediction network does seem to capture the factors like scene motion and visibility well (see Sec.~\ref{sec:exp_vis}), and could potentially be more important for other more challenging datasets.



\begin{figure}[t]
    \centering
    \includegraphics[width=\linewidth]{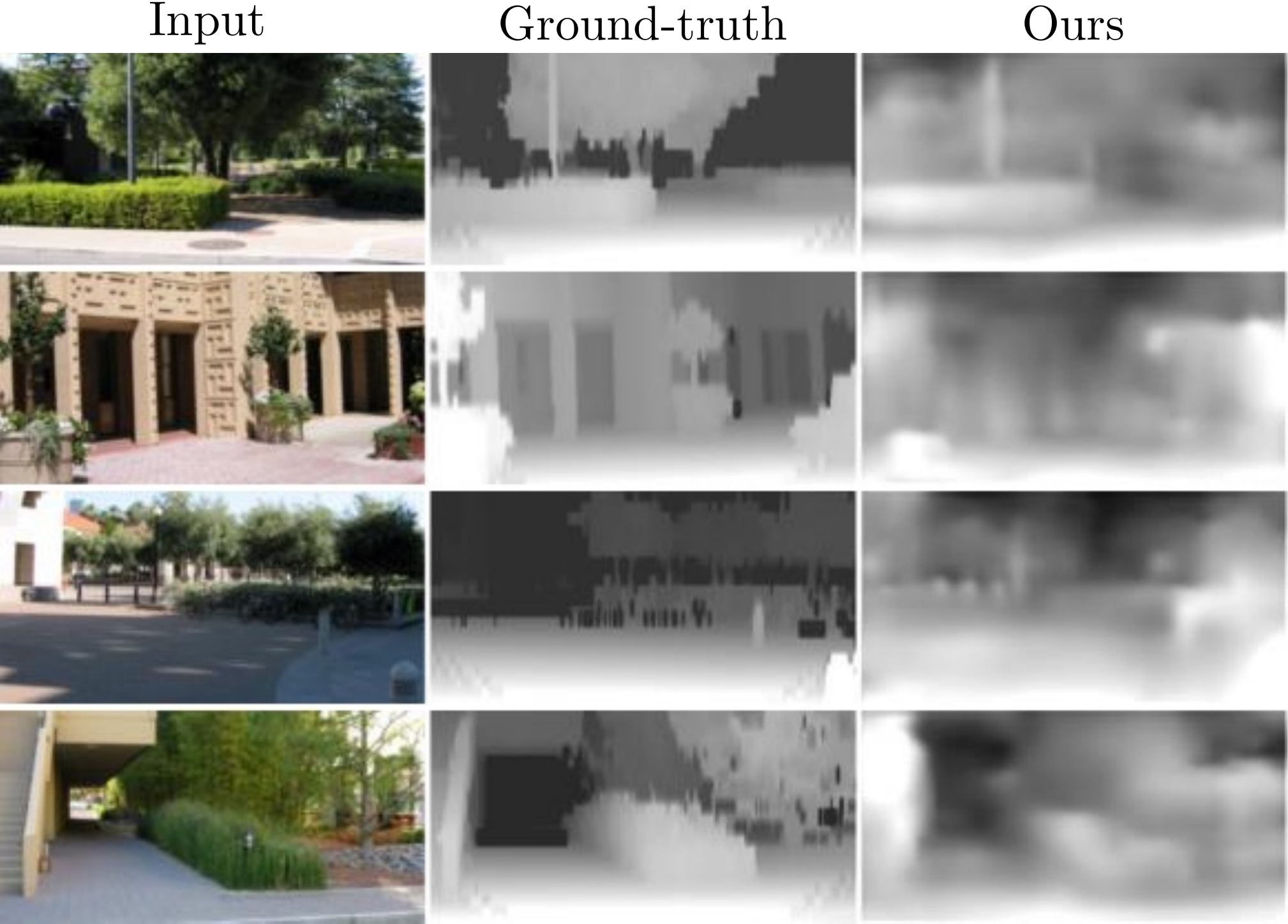}
    \caption{Our sample predictions on the Make3D dataset. Note that our model is trained on KITTI + Cityscapes only, and directly tested on Make3D.}
    \label{fig:make3d}
\end{figure}

\paragraph{Make3D}
To evaluate the generalization ability of our single-view depth model, we directly apply our model trained on Cityscapes + KITTI to the Make3D dataset unseen during training. While there still remains a significant performance gap between our method and others supervised using Make3D ground-truth depth (see Table~\ref{tab:make3d}), our predictions are able to capture the global scene layout reasonably well without any training on the Make3D images (see Figure~\ref{fig:make3d}).

\begin{table}[t]
  \centering
  \resizebox{0.99\linewidth}{!}{
 \begin{tabular}{lcccccc}
 \toprule
 Method & \multicolumn{2}{c}{Supervision} & \multicolumn{4}{c}{Error metric} \\  
 \cmidrule(lr){2-3}
 \cmidrule(lr){4-7}
  & Depth & Pose & Abs Rel & Sq Rel & RMSE & RMSE log \tabularnewline
 \midrule
 Train set mean & \checkmark &  & 0.876 & 13.98  & 12.27 & 0.307 \tabularnewline
 Karsch \etal~\cite{karsch2014depth}& \checkmark &  & 0.428 & 5.079  & 8.389 & 0.149 \tabularnewline
 Liu \etal~\cite{liu2014discrete} & \checkmark & & 0.475 & 6.562  & 10.05 & 0.165 \tabularnewline
 Laina \etal~\cite{laina2016deeper} & \checkmark & & 0.204 & 1.840  & 5.683 & 0.084 \tabularnewline
 Godard \etal~\cite{godard2016unsupervised}&  & \checkmark  & 0.544 & 10.94 & 11.76 & 0.193 \tabularnewline
 \midrule
 \textbf{Ours} & & & 0.383 & 5.321 & 10.47 & 0.478 \\
  \bottomrule
\end{tabular}
}

\vspace{10pt}
  \caption{Results on the Make3D dataset \cite{saxena2009make3d}. Similar to ours, Godard~\etal~\cite{godard2016unsupervised} do not utilize any of the Make3D data during training, and directly apply the model trained on KITTI+Cityscapes to the test set. Following the evaluation protocol of~\cite{godard2016unsupervised}, the errors are only computed where depth is less than $70$ meters in a central image crop.}
    \label{tab:make3d}
\end{table}

 \subsection{Pose estimation}
 \label{sec:odom}
\begin{figure}[t]
    \centering
    \includegraphics[scale=0.36]{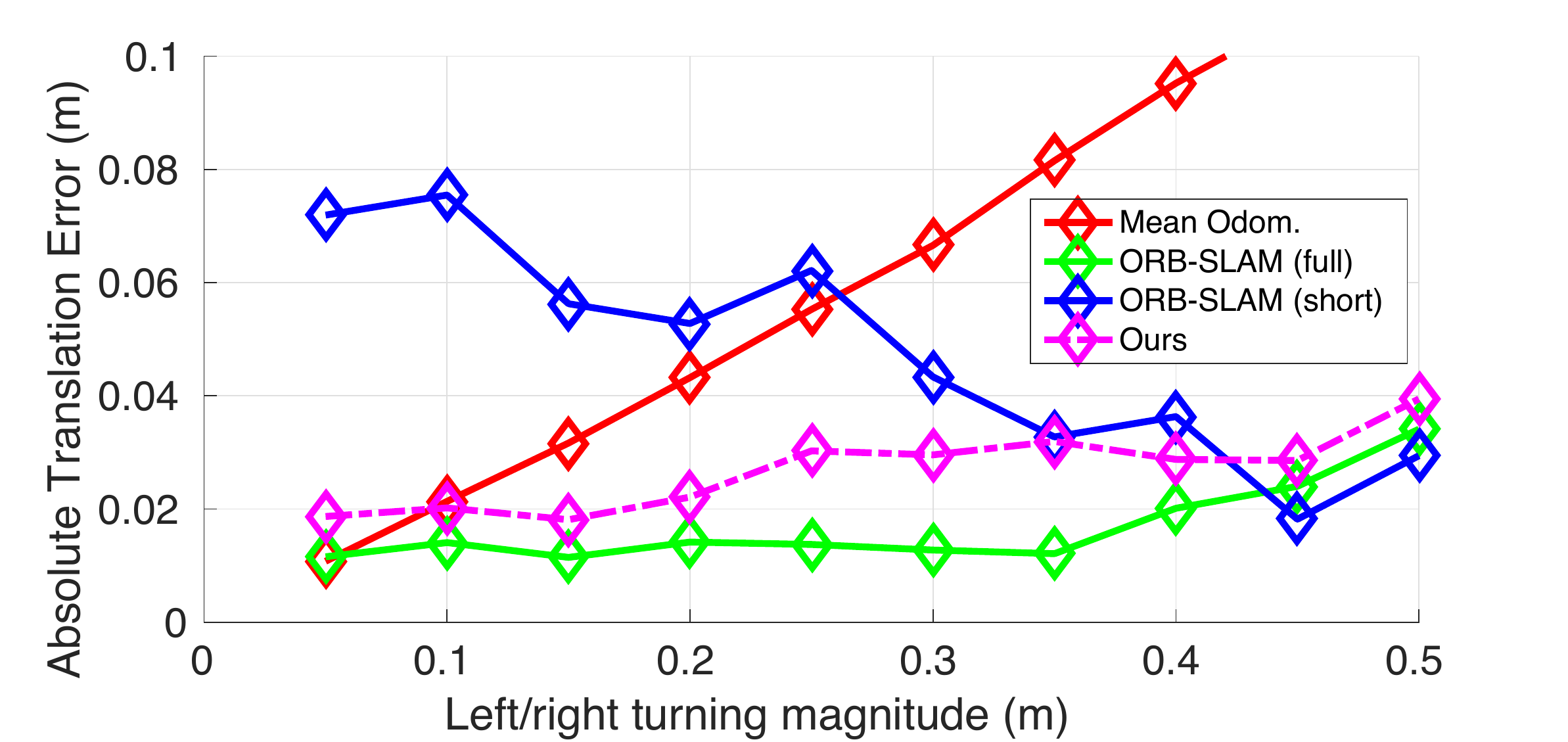}
    \caption{Absolute Trajectory Error (ATE) at different left/right turning magnitude (coordinate difference in the side-direction between the start and ending frame of a testing sequence). Our method performs significantly better than \texttt{ORB-SLAM (short)} when side rotation is small, and is comparable with \texttt{ORB-SLAM (full)} across the entire spectrum.}
    \label{fig:ate}
\end{figure}

To evaluate the performance of our pose estimation network, we applied our system to the official KITTI odometry split (containing $11$ driving sequences with ground truth odometry obtained through the IMU/GPS readings, which we use for evaluation purpose only), and used sequences $00$-$08$ for training and $09$-$10$ for testing. In this experiment, we fix the length of input image sequences to our system to $5$ frames. We compare our ego-motion estimation with two variants of monocular ORB-SLAM~\cite{mur2015orb} (a well-established SLAM system): 1) \texttt{ORB-SLAM (full)}, which recovers odometry using all frames of the driving sequence (i.e.\ allowing loop closure and re-localization), and 2) \texttt{ORB-SLAM (short)}, which runs on $5$-frame snippets (same as our input setting). Another baseline we compare with is the dataset mean of car motion (using ground-truth odometry) for $5$-frame snippets. To resolve scale ambiguity during evaluation, we first optimize the scaling factor for the predictions made by each method to best align with the ground truth, and then measure the Absolute Trajectory Error (ATE)~\cite{mur2015orb} as the metric. ATE is computed on $5$-frame snippets and averaged over the full sequence.\footnote{For evaluating \texttt{ORB-SLAM (full)} we break down the trajectory of the full sequence into $5$-frame snippets with the reference coordinate frame adjusted to the central frame of each snippet.} As shown in Table~\ref{tab:ate} and Fig.~\ref{fig:ate}, our method outperforms both baselines (mean odometry and \texttt{ORB-SLAM (short)}) that share the same input setting as ours, but falls short of \texttt{ORB-SLAM (full)}, which leverages whole sequences ($1591$ for seq.\ $09$ and $1201$ for seq.\ $10$) for loop closure and re-localization. 

For better understanding of our pose estimation results, we show in Figure~\ref{fig:ate} the ATE curve with varying amount of side-rotation by the car between the beginning and the end of a sequence. Figure~\ref{fig:ate} suggests that our method is significantly better than \texttt{ORB-SLAM (short)} when the side-rotation is small (i.e. car mostly driving forward), and comparable to \texttt{ORB-SLAM (full)} across the entire spectrum. The large performance gap between ours and \texttt{ORB-SLAM (short)} suggests that our learned ego-motion could potentially be used as an alternative to the local estimation modules in monocular SLAM systems.

\begin{table}[t]
\centering
\scalebox{0.88}{
\begin{tabular}{lcc}
\toprule
\textbf{Method} & \textbf{Seq. $09$} & \textbf{Seq. $10$}
\tabularnewline
\midrule
\textbf{ORB-SLAM (full)} & $\mathbf{0.014 \pm 0.008}$ & $\mathbf{0.012 \pm 0.011}$  \tabularnewline
\midrule
\textbf{ORB-SLAM (short)} & $0.064 \pm 0.141$ & $0.064 \pm 0.130$ \tabularnewline
\textbf{Mean Odom.} & $0.032 \pm 0.026$ & $0.028 \pm 0.023$ \tabularnewline
\textbf{Ours} & $\mathbf{0.021 \pm 0.017}$ & $\mathbf{0.020 \pm 0.015}$ \tabularnewline
\bottomrule
\end{tabular}}
\vspace{0.1cm}
\caption{Absolute Trajectory Error (ATE) on the KITTI odometry split averaged over all $5$-frame snippets (lower is better). Our method outperforms baselines with the same input setting, but falls short of \texttt{ORB-SLAM (full)} that uses strictly more data.}
\label{tab:ate}
\end{table}

\vspace{-0.04in}

\subsection{Visualizing the explainability prediction}
\label{sec:exp_vis}
\begin{figure}
    \centering
    \includegraphics[width=\linewidth]{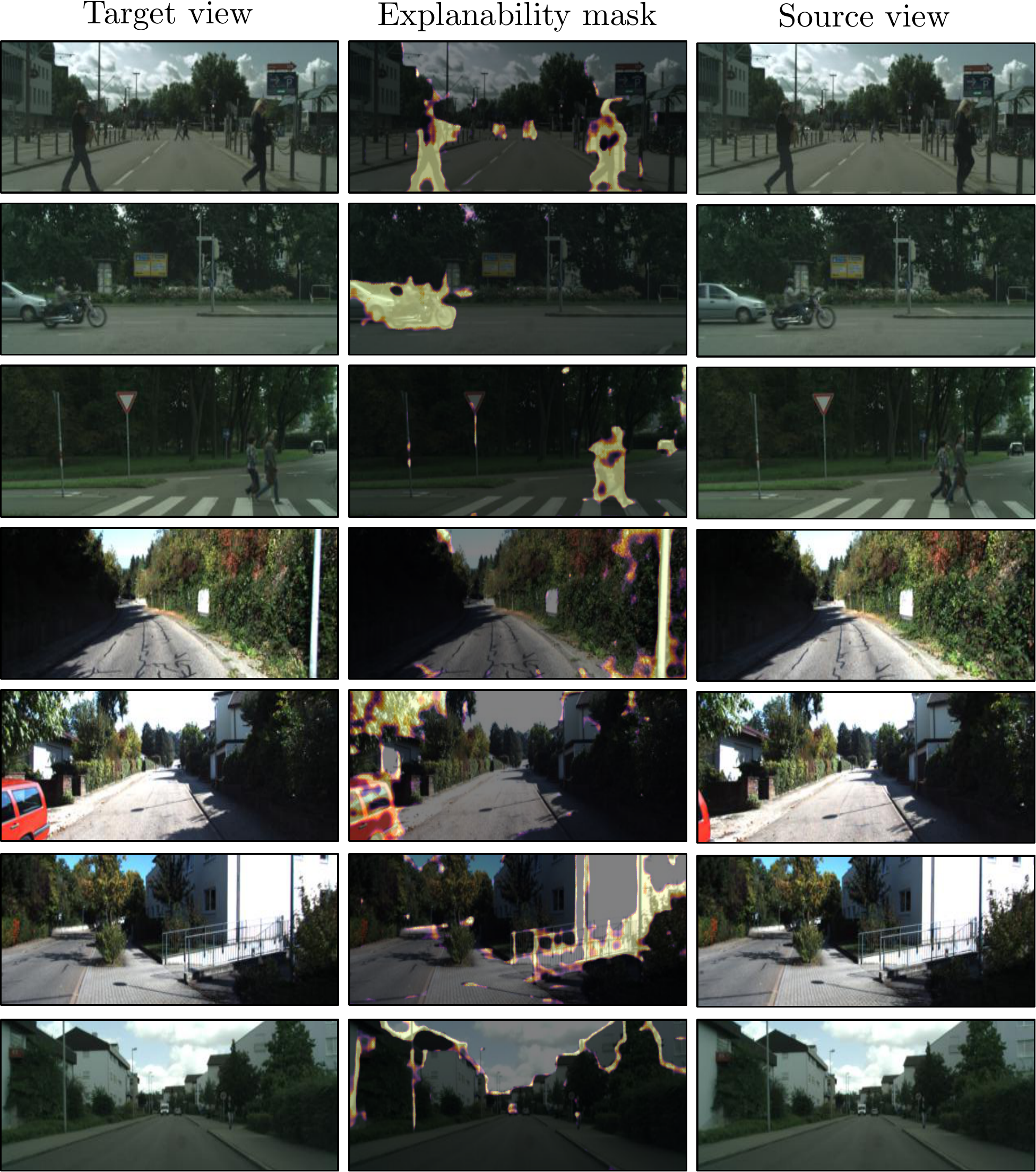}
    \caption{Sample visualizations of the explainability masks. Highlighted pixels are predicted to be unexplainable by the network due to motion (rows 1--3), occlusion/visibility (rows 4--5), or other factors (rows 7--8). }
    \label{fig:exp}
\end{figure}

We visualize example explainability masks predicted by our network in Figure~\ref{fig:exp}. The first three rows suggest that the network has learned to identify dynamic objects in the scene as unexplainable by our model, and similarly, rows 4--5 are examples of objects that disappear from the frame in subsequent views. The last two rows demonstrate the potential downside of explainability-weighted loss: the depth CNN has low confidence in predicting thin structures well, and tends to mask them as unexplainable. 

\vspace{-0.1in}

\section{Discussion}
We have presented an end-to-end learning pipeline that utilizes the task of view synthesis for supervision of single-view depth and camera pose estimation. The system is trained on unlabeled videos, and yet performs comparably with approaches that require ground-truth depth or pose for training. Despite good performance on the benchmark evaluation, our method is by no means close to solving the general problem of unsupervised learning of 3D scene structure inference. A number of major challenges are yet to be addressed: 1) our current framework does not explicitly estimate scene dynamics and occlusions (although they are implicitly taken into account by the explainability masks), both of which are critical factors in 3D scene understanding. Direct modeling of scene dynamics through motion segmentation (e.g.~\cite{Vijayanarasimhan2017SfM,ranftl2016dense}) could be a potential solution;  2) our framework assumes the camera intrinsics are given, which forbids the use of random Internet videos with unknown camera types/calibration -- we plan to address this in future work; 3) depth maps are a simplified representation of the underlying 3D scene. It would be interesting to extend our framework to learn full 3D volumetric representations (e.g. ~\cite{tulsiani2017multi}).

Another interesting area for future work would be to investigate in more detail the representation learned by our system. In particular, the pose network likely uses some form of image correspondence in estimating the camera motion, whereas the depth estimation network likely recognizes common structural features of scenes and objects. It would be interesting to probe these, and investigate the extent to which our network already performs, or could be re-purposed to perform, tasks such as object detection and semantic segmentation.

\vspace{-0.1in}

{\footnotesize
\paragraph{Acknowledgments:} We thank our colleagues, Sudheendra Vijayanarasimhan, Susanna Ricco, Cordelia Schmid, Rahul Sukthankar, and Katerina Fragkiadaki for their help. We also thank the anonymous reviewers for their valuable comments. TZ would like to thank Shubham Tulsiani for helpful discussions, and Clement Godard for sharing the evaluation code. This work is also partially funded by Intel/NSF VEC award IIS-1539099. 
}

{\small
\bibliographystyle{ieee}
\bibliography{egbib}
}

\end{document}